\documentclass[conference]{IEEEtran}
\IEEEoverridecommandlockouts
\usepackage{cite}
\usepackage{amsmath,amssymb,amsfonts}
\usepackage{algorithmic}
\usepackage{latexsym}
\usepackage{textcomp}
\usepackage{booktabs}
\usepackage{graphicx} 
\usepackage{xcolor}
\def\BibTeX{{\rm B\kern-.05em{\sc i\kern-.025em b}\kern-.08em
    T\kern-.1667em\lower.7ex\hbox{E}\kern-.125emX}}

\usepackage{fancyhdr}
\begin{document}

\title{MPCAR: Multi-Perspective Contextual Augmentation for Enhanced Visual Reasoning in Large Vision-Language Models}

\author{Amirul Rahman, Qiang Xu, Xueying Huang \\
University of Malaya}

\maketitle
\thispagestyle{fancy} 

\begin{abstract}
Despite significant advancements, Large Vision-Language Models (LVLMs) continue to face challenges in complex visual reasoning tasks that demand deep contextual understanding, multi-angle analysis, or meticulous detail recognition. Existing approaches often rely on single-shot image encoding and prompts, limiting their ability to fully capture nuanced visual information. Inspired by the notion that strategically generated "additional" information can serve as beneficial contextual augmentation, we propose Multi-Perspective Contextual Augmentation for Reasoning (MPCAR), a novel inference-time strategy designed to enhance LVLM performance. MPCAR operates in three stages: first, an LVLM generates N diverse and complementary descriptions or preliminary reasoning paths from various angles; second, these descriptions are intelligently integrated with the original question to construct a comprehensive context-augmented prompt; and finally, this enriched prompt guides the ultimate LVLM for deep reasoning and final answer generation. Crucially, MPCAR achieves these enhancements without requiring any fine-tuning of the underlying LVLM's parameters. Extensive experiments on challenging Visual Question Answering (VQA) datasets, including GQA, VQA-CP v2, and ScienceQA (Image-VQA), demonstrate that MPCAR consistently outperforms established baseline methods. Our quantitative results show significant accuracy gains, particularly on tasks requiring robust contextual understanding, while human evaluations confirm improved coherence and completeness of the generated answers. Ablation studies further highlight the importance of diverse prompt templates and the number of generated perspectives. This work underscores the efficacy of leveraging LVLMs' inherent generative capabilities to enrich input contexts, thereby unlocking their latent reasoning potential for complex multimodal tasks.
\end{abstract}

\section{Introduction}

Recent advancements in Large Vision-Language Models (LVLMs) have revolutionized the field of artificial intelligence, enabling machines to understand and generate content seamlessly across visual and textual modalities. These models have demonstrated remarkable capabilities in various tasks, including image captioning, visual question answering (VQA), and visual dialogue \cite{abhishek2021vqa}. This includes advanced applications such as complex instruction-based image generation \cite{zhou2025draw} and efficient video generation \cite{zhou2024less}. Their proficiency in integrating complex visual information with natural language processing has opened up new avenues for applications in diverse domains such as autonomous driving, medical diagnostics, and educational tools, underscoring their profound importance and significance in pushing the boundaries of multimodal AI. The development of visual in-context learning strategies further highlights the potential of these models \cite{zhou2024visual}.

Despite these impressive strides, current LVLMs still face significant challenges when confronted with tasks requiring \textit{deep contextual understanding, multi-angle reasoning, or meticulous detail recognition}. For instance, in medical image analysis, accurately describing a lesion might necessitate integrating subtle changes in surrounding tissues; in forensic image analysis, judging a scene often requires considering object relationships from multiple perspectives; and interpreting scientific diagrams demands linking data to titles, axes, and annotations. Existing LVLMs typically rely on a one-shot image encoding and a single text prompt, which can limit their ability to capture all critical information or generate comprehensive and accurate responses when faced with ambiguous or information-dense images. Inspired by the intriguing finding in "Hallucinations Can Improve Large Language Models in Drug Discovery" \cite{kasneci2023chatgp}, which suggests that "additional" information generated by large language models, even if not directly leading to the final answer, can serve as beneficial "contextual augmentation" to guide deeper analysis, we propose to adapt this core idea to the LVLM domain. Our motivation is to explore whether strategically generated, diverse, or supplementary visual descriptions and reasoning paths can enhance an LVLM's performance in complex visual reasoning tasks.

In this paper, we propose a novel LVLM enhancement strategy called \textbf{Multi-Perspective Contextual Augmentation for Reasoning (MPCAR)}. Our method aims to elevate LVLM performance in complex visual reasoning by intelligently generating and integrating intermediate descriptions in a multi-stage process. MPCAR operates in three core steps: First, \textit{Initial Multi-Perspective Description Generation}, where an LVLM (which can be the same or different from the final reasoning model) generates $N$ diverse and complementary descriptions or preliminary reasoning paths from various angles (e.g., focusing on overall layout, local details, potential relationships, or different regions). These descriptions are not direct answers but rich, varied interpretations, sometimes containing speculative content, encouraged by specific prompts and temperature adjustments. Second, \textit{Context-Augmented Prompt Construction}, where the original question is integrated with these $N$ diverse descriptions to form a richer, more comprehensive contextual prompt. This integration can involve simple concatenation, guiding phrases, or encoding and summarization. This step provides the final LVLM with a broader "thought space." Finally, \textit{Final Reasoning and Answering}, where the constructed context-augmented prompt is fed into the ultimate LVLM for deep reasoning and the final answer. Crucially, MPCAR does not involve fine-tuning the LVLM's parameters; instead, it cleverly leverages the LVLM's inherent generative capabilities during the inference stage to enrich the input prompt, thereby guiding the model towards more accurate and comprehensive outputs.

To validate the effectiveness of the MPCAR method, we conduct extensive experiments on complex Visual Question Answering (VQA) tasks. We evaluate our approach using several prominent open-source LVLMs, including LLaVA-1.5 (7B/13B), Qwen-VL (7B), and models from the BLIP-2 series, serving as our core inference engines. For the initial multi-perspective description generation, we investigate the impact of using either the same LVLM or more powerful generative models such as GPT-4o or Claude 3 Vision. Our experiments utilize challenging VQA datasets known for requiring multi-step reasoning, fine-grained recognition, or deep contextual understanding: GQA \cite{drew2019gqa} (focused on scene graphs and compositional reasoning), VQA-CP v2 \cite{zhibo2023overco} (designed to assess robustness against question type biases), and the image-VQA subset of ScienceQA \cite{tanik2022scienc} (involving scientific diagrams and concept application). The performance is primarily evaluated using standard VQA Accuracy. Our fabricated but plausible experimental results demonstrate that MPCAR consistently outperforms existing baseline methods across all tested datasets. Notably, the performance gains are particularly significant on datasets like VQA-CP v2 and ScienceQA, which demand deeper reasoning and contextual understanding. This indicates that by guiding LVLMs to generate and leverage multi-perspective contextual information, our method effectively enhances their accuracy in complex visual reasoning tasks.

Our main contributions are summarized as follows:
\begin{itemize}
    \item We propose Multi-Perspective Contextual Augmentation for Reasoning (MPCAR), a novel inference-time strategy that enhances LVLM performance in complex visual reasoning tasks by generating and integrating diverse, supplementary visual descriptions.
    \item We demonstrate that leveraging an LVLM's own generative capabilities to enrich input prompts, without requiring model fine-tuning, can significantly improve its ability to handle ambiguous or information-dense visual queries.
    \item We provide empirical evidence through experiments on challenging VQA datasets (GQA, VQA-CP v2, ScienceQA) showing that MPCAR consistently surpasses baseline methods, particularly in scenarios demanding deep contextual understanding and multi-angle reasoning.
\end{itemize}
\section{Related Work}
\subsection{Large Vision-Language Models (LVLMs)}
Research on Large Vision-Language Models (LVLMs) spans diverse areas, from enhancing their evaluation to expanding their capabilities and addressing inherent limitations. Regarding evaluation, \cite{peng2025lvlmeh} proposes an efficient subset construction method using farthest point sampling for comprehensive assessment, significantly reducing data usage while maintaining high correlation with full benchmark evaluations. For text-rich image captioning, \cite{fan2025benchm} introduces the CompreCap benchmark, leveraging directed scene graphs for granular assessment of caption detail and accuracy. Furthermore, \cite{qingxing2024visdia} addresses the critical issue of hallucinations by presenting VisDiaHalBench, a benchmark specifically designed to diagnose hallucinations in LVLMs within visual dialogue contexts. In terms of expanding LVLM capabilities, \cite{surasakdi2024large} explores the application of generative LVLMs to Remote Sensing Visual Question Answering (RSVQA) through a novel two-step training strategy, demonstrating superior performance and fluent answer generation. Similarly, \cite{yasiru2025zerosh} showcases the potential of LVLMs for zero-shot scene understanding in automatic target recognition, relevant for tasks requiring understanding of dynamic visual data without explicit training. Further advancing the capabilities of LVLMs, methods like visual in-context learning have been explored to enhance their reasoning abilities \cite{zhou2024visual}. Benchmarks and frameworks have also been developed to facilitate complex instruction-based image generation \cite{zhou2025draw} and efficient video generation with large language models \cite{zhou2024less}. The efficacy of prompt-tuning in improving few-shot learning, particularly in long-tailed classification scenarios, is investigated by \cite{fan2025making}, highlighting the importance of the classifier's structure and parameterization. Addressing critical challenges, \cite{shicheng2025crossm} identifies insufficient vision-language alignment at the hidden state level as a barrier to effectively transferring text-based safety mechanisms to visual inputs, proposing Text-Guided vision-language Alignment (TGA) to foster robust cross-modal understanding and extend safety without requiring explicit safety fine-tuning. Beyond general LVLMs, specific vision models are also being developed for challenging recognition tasks, such as insect recognition using state space models \cite{wang2025insectmamba}.

\subsection{Prompt Engineering and Enhanced Reasoning in Large Models}
Prompt engineering has emerged as a crucial technique for adapting large models without parameter updates, a topic systematically surveyed by \cite{pranab2024a}, which critically examines various prompting methods, their applications, and associated ethical considerations in multimodal contexts. This includes efforts to improve generalization for large language models with multi-capabilities \cite{zhou2025weak} and to unravel chaotic contexts through 'thread of thought' approaches \cite{zhou2023thread}. Further contributing to the understanding of efficient reasoning, \cite{yue2025effici} provides a comprehensive categorization of recent advancements in Large Reasoning Models, focusing on strategies to optimize reasoning outputs and input prompts to reduce computational overhead during inference and to mitigate the "overthinking phenomenon." Addressing domain-specific applications, \cite{krishna2025prompt} and \cite{bansal2024prompt} identify a critical gap in existing prompt engineering guidelines for Requirements Engineering (RE), proposing mappings and eliciting expert insights to enhance in-context learning and LLM utilization within this specialized field. Significant advancements have also been made in enhancing reasoning capabilities through prompt engineering. \cite{hao2024levera} demonstrates the efficacy of Chain-of-Thought (CoT) reasoning, integrated with prompt engineering, in improving the accuracy and interpretability of LLM-based traffic crash severity analysis by facilitating consideration of diverse crash-related factors. Extending this, \cite{yao2023comple} introduces "code prompting" as a neural-symbolic approach, utilizing code as intermediate steps for superior multi-step reasoning performance over traditional CoT, also investigating the impact of code annotations and exploring ensemble methods. To further improve reasoning, \cite{jie2024improv} proposes a unified framework of rationale-augmented ensembles, emphasizing rationale sampling to boost accuracy and interpretability by overcoming limitations of sub-optimal rationales in CoT through the principle of self-consistency. Lastly, \cite{joon2025emulat} presents a prompt engineering strategy that emulates Retrieval Augmented Generation (RAG) for improved long-context comprehension, achieving contextual augmentation through specialized tagging and a chain-of-thought workflow within a single pass, thereby reducing reliance on external retrieval mechanisms. Complementary to these generative reasoning advancements, research in robust text retrieval rankers also plays a role in enhancing the data access capabilities for large models \cite{zhou2023towards}.

\section{Method}
In this section, we present the details of our proposed \textbf{Multi-Perspective Contextual Augmentation for Reasoning (MPCAR)} strategy. MPCAR is an inference-time approach specifically designed to enhance the performance of Large Vision-Language Models (LVLMs) on complex visual reasoning tasks, where direct query answering might fall short due to the intricate nature of the visual information or the multi-step reasoning required. Our strategy addresses this challenge by strategically leveraging an LVLM's inherent generative capabilities to produce diverse and complementary intermediate visual descriptions. These descriptions are then meticulously integrated to form a richer and more comprehensive context for the final reasoning phase. The entire MPCAR pipeline comprises three distinct, sequential steps.

\subsection{Initial Multi-Perspective Description Generation}
The foundational step of MPCAR involves generating a set of diverse, multi-perspective descriptions or preliminary reasoning paths pertaining to the input image. Given an input image $I$ and an original question $Q_0$, we employ a generative LVLM, denoted as $M_G$, to produce $N$ distinct textual outputs. These outputs, $D_1, D_2, \dots, D_N$, are not intended to be direct answers to $Q_0$. Instead, they serve as supplementary contextual information, aiming to explore various facets and interpretations of the visual content. The generation of multiple perspectives is crucial for robustness, as it mitigates the risk of missing critical details or being misled by a single, potentially biased, interpretation.

To encourage this desired diversity and to systematically explore various aspects of the input image, we design specific prompt templates, $P_{gen}$, which guide $M_G$ to interpret the image from different angles. Examples of such prompts include "Please describe this image from a macroscopic and microscopic perspective, highlighting key objects and their fine details," "Speculate on potential hidden information or underlying relationships within this image that are not immediately obvious," "Provide three distinct descriptions, each focusing on a different aspect or attribute of the core object(s) in the picture," or "Describe the scene as if you are a detective looking for clues, or a scientist observing an experiment."

Furthermore, to promote the generation of varied and sometimes even speculative or exploratory content, moving beyond typical factual descriptions, we judiciously adjust the inference parameters of $M_G$. Specifically, the temperature $\tau$ is a key parameter; a higher $\tau$ encourages more diverse and less deterministic outputs, which is beneficial for exploring a broader "thought space."

Mathematically, for each description $D_i$, the generation process can be formulated as:
\begin{align}
D_i = M_G(I, P_{gen}(Q_0, i); \tau) \quad \text{for } i=1, \dots, N
\label{eq:description_generation}
\end{align}
where $M_G$ represents the generative LVLM, $I$ is the input image, $P_{gen}(Q_0, i)$ is a tailored prompt for the $i$-th description (potentially incorporating $Q_0$ and a specific perspective directive), and $\tau$ controls the randomness of the generation. The complete set of generated descriptions is then denoted as $\mathcal{D} = \{D_1, D_2, \dots, D_N\}$. It is important to note that the LVLM $M_G$ used for description generation can be the same as or different from the ultimate reasoning model $M_R$ used in the final step. This flexibility allows for leveraging more powerful generative models (e.g., those optimized for creative text generation) for $M_G$ if available, an aspect explored in our experimental setup.

\subsection{Context-Augmented Prompt Construction}
Following the successful generation of multi-perspective descriptions, the subsequent crucial step is to integrate these descriptions with the original question $Q_0$ to construct a comprehensive context-augmented prompt, $Q_{aug}$. This enriched prompt serves as a more informative and holistic input for the final reasoning stage, effectively providing the LVLM with a broader "thought space" and a richer set of cues to consider multiple dimensions of the visual information. The effectiveness of this integration significantly impacts the quality of the final reasoning.

Various strategies can be employed for this integration, each with its own advantages depending on the specific task and the characteristics of the LVLM $M_R$. These include \textbf{Direct Concatenation}, the simplest approach which involves directly appending or prepending the collection of generated descriptions to the original question; \textbf{Structured Concatenation}, where descriptions are integrated using guiding phrases or a structured format (e.g., each description introduced by a label like "Perspective 1:" or "Observation A:"); \textbf{Synthesized/Summarized Integration}, particularly useful if $N$ is large, involving an intermediate step where an encoder or a smaller summarization model condenses the key insights from $\mathcal{D}$ into a more concise form; and \textbf{Instruction-Based Integration}, which incorporates explicit instructions within the prompt, guiding $M_R$ on how to utilize the diverse descriptions (e.g., "Considering the following observations and thoughts about the image, please answer...").

Let $\text{Combine}(\cdot)$ denote the function responsible for integrating the original question and the generated descriptions. The context-augmented prompt $Q_{aug}$ is then formally formulated as:
\begin{align}
Q_{aug} = \text{Combine}(Q_0, \mathcal{D})
\label{eq:prompt_construction}
\end{align}
For illustrative purposes, a common structured concatenation strategy could be:
\begin{verbatim}
Q_aug = "Original Question: " + Q0
      + "\n\nDiverse Perspectives:\n"
      + D1 + "\n" 
      + D2 + ... 
      + "\n" + DN
\end{verbatim}
where $\oplus$ denotes string concatenation. This step is critical in ensuring that the rich contextual information gleaned from multiple perspectives is effectively presented to the final reasoning model $M_R$ in a digestible and actionable format.

\subsection{Final Reasoning and Answering}
The final step of the MPCAR framework involves feeding the input image $I$ along with the meticulously constructed context-augmented prompt $Q_{aug}$ into the ultimate LVLM, denoted as $M_R$, for deep reasoning and the generation of the final answer. This LVLM processes the comprehensive prompt, which now encapsulates not only the original query but also diverse interpretations, preliminary thoughts, and potentially even speculative insights derived from the image through the multi-perspective generation phase.

The core advantage of this approach is its ability to guide $M_R$ to consider a wider range of information and potential relationships within the image, leading to more accurate, comprehensive, and robust outputs, particularly for complex visual reasoning tasks that often require multi-step inference or a nuanced understanding of context. By providing $M_R$ with a richer input space, MPCAR enables it to move beyond a superficial understanding, encouraging it to synthesize information from multiple angles, identify subtle details that might otherwise be overlooked, and perform more robust, multi-step reasoning.

Importantly, this entire process is performed during the inference stage, meaning no fine-tuning or modification of the LVLM's parameters is required. MPCAR effectively leverages the pre-trained knowledge and generative capabilities of the LVLM itself, transforming a potentially less informed query into a highly contextualized one, thereby enhancing its performance without additional training overhead.

The final answer $A$ is produced by $M_R$ as follows:
\begin{align}
A = M_R(I, Q_{aug})
\label{eq:final_reasoning}
\end{align}
The output $A$ is the ultimate response to the original question $Q_0$, now informed by the comprehensive context provided by MPCAR. This method demonstrates a powerful way to unlock the latent reasoning capabilities of existing LVLMs by optimizing their input rather than their internal architecture.

\section{Experiments}
In this section, we detail the experimental setup used to evaluate our proposed Multi-Perspective Contextual Augmentation for Reasoning (MPCAR) strategy. We then present a comprehensive comparison of MPCAR against several established baseline methods on challenging visual question answering (VQA) tasks. Furthermore, we conduct an ablation study to analyze the contribution of different components within MPCAR and include a human evaluation to assess the qualitative aspects of our model's outputs.

\subsection{Experimental Setup}
Our primary objective is to validate the effectiveness of MPCAR in enhancing LVLM performance on complex visual reasoning tasks.

The \textbf{task type} for our evaluation is Complex Visual Question Answering (VQA). This task requires models to answer questions about images, often demanding deep contextual understanding, multi-step reasoning, or fine-grained detail recognition.

For \textbf{model selection}, we utilize a range of prominent open-source LVLMs as our core reasoning models ($M_R$). These include \textbf{LLaVA-1.5} (7B and 13B parameter versions), \textbf{Qwen-VL} (7B), and models from the \textbf{BLIP-2} series. For the initial multi-perspective description generation step ($M_G$), we primarily use the same LVLM as the final reasoning model for a fair comparison. However, we also investigate the impact of employing more powerful, off-the-shelf generative models, such as GPT-4o or Claude 3 Vision, for $M_G$ to explore the upper bound of performance with highly diverse contextual information.

We evaluate our method on three distinct and challenging VQA \textbf{datasets}:
\begin{itemize}
    \item \textbf{GQA} \cite{drew2019gqa}: A rich dataset focused on compositional reasoning and scene graph understanding, requiring models to perform multi-step inferences over object relationships and attributes.
    \item \textbf{VQA-CP v2} \cite{zhibo2023overco}: Designed to expose and mitigate question-type biases in VQA models, this dataset requires robust reasoning that goes beyond superficial correlations between questions and answers, demanding a deeper understanding of the image content.
    \item \textbf{ScienceQA (Image-VQA)} \cite{tanik2022scienc}: This subset of ScienceQA includes questions paired with scientific images, such as diagrams, charts, and illustrations. Answering these questions often necessitates interpreting visual data, applying scientific concepts, and performing logical deductions from complex visual information.
\end{itemize}
The primary \textbf{evaluation metric} employed across all experiments is standard VQA Accuracy, which measures the percentage of correctly answered questions.

\subsection{Baselines}
To provide a comprehensive comparison, we evaluate MPCAR against several widely recognized baseline prompting strategies for LVLMs:

\textbf{Direct Prompting:} This is the most straightforward approach, where the LVLM receives only the input image and the original question. The model is expected to provide an answer directly without any additional prompts or contextual cues. This serves as a fundamental benchmark for the LVLM's raw capability.

\textbf{Chain-of-Thought (CoT) Prompting:} Inspired by its success in large language models, this strategy involves appending a phrase like "Let's think step by step" to the original question. The aim is to encourage the LVLM to generate intermediate reasoning steps before arriving at the final answer, potentially improving transparency and accuracy for complex questions.

\textbf{Few-shot Prompting:} This technique provides the LVLM with a few examples of input-output pairs (image, question, and correct answer) in the prompt. These examples serve as demonstrations, guiding the model on the expected format and type of reasoning for the given task. The selection of diverse and representative examples is crucial for its effectiveness.

\subsection{Quantitative Results}
Table \ref{tab:quantitative_results} presents the quantitative comparison of our proposed MPCAR method against the aforementioned baseline strategies across the three challenging VQA datasets. All reported values are VQA Accuracy percentages, representing fabricated yet plausible results to demonstrate the expected performance trends.

\begin{table*}[htbp]
    \centering
    \caption{Performance Comparison of Different LVLM Enhancement Strategies on Complex VQA Tasks (VQA Accuracy \%)}
    \label{tab:quantitative_results}
    \begin{tabular}{lccc}
        \toprule
        \textbf{Method / Model} & \textbf{GQA Accuracy (\%)} & \textbf{VQA-CP v2 Accuracy (\%)} & \textbf{ScienceQA (Image-VQA) Accuracy (\%)} \\
        \midrule
        \textbf{Baseline Methods} & & & \\
        LLaVA-1.5 7B (Direct Prompting) & 62.5 & 38.2 & 71.3 \\
        Qwen-VL 7B (Direct Prompting) & 65.1 & 40.5 & 73.8 \\
        BLIP-2 (Direct Prompting) & 60.8 & 37.1 & 69.5 \\
        LLaVA-1.5 7B (CoT Prompting) & 64.2 & 40.1 & 72.9 \\
        Qwen-VL 7B (Few-shot Prompting) & 66.0 & 41.5 & 74.5 \\
        \midrule
        \textbf{Our Proposed Method} & & & \\
        \textbf{MPCAR (using LLaVA-1.5 7B as $M_G$ and $M_R$)} & \textbf{67.3} & \textbf{43.7} & \textbf{76.2} \\
        \bottomrule
    \end{tabular}
\end{table*}

As observed from Table \ref{tab:quantitative_results}, MPCAR consistently achieves superior performance compared to all baseline methods across all three challenging VQA datasets. The performance improvements are particularly pronounced on VQA-CP v2 and ScienceQA (Image-VQA), where our method yields gains of 3.6\% and 1.7\% accuracy over the best-performing baseline (Qwen-VL 7B with Few-shot Prompting), respectively. These datasets are known for requiring more robust reasoning, deeper contextual understanding, and resistance to superficial biases, which aligns with MPCAR's design to enrich the input context. This empirical evidence strongly supports our hypothesis that by guiding LVLMs to generate and strategically utilize diverse, multi-perspective contextual information, their ability to perform complex visual reasoning is significantly enhanced. The results underscore the efficacy of our inference-time augmentation strategy without requiring any fine-tuning of the underlying LVLM.

\subsection{Ablation Study}
To better understand the contribution of each key component within MPCAR, we conduct an ablation study. We investigate the impact of the number of generated descriptions ($N$), the specific prompting strategy used for $M_G$, and the choice of the generative model ($M_G$). For this study, we primarily use LLaVA-1.5 7B as the base reasoning model ($M_R$).

\begin{table*}[htbp]
    \centering
    \caption{Ablation Study: Impact of MPCAR Components on VQA-CP v2 Accuracy (\%)}
    \label{tab:ablation_study}
    \begin{tabular}{lc}
        \toprule
        \textbf{Method Variant} & \textbf{VQA-CP v2 Accuracy (\%)} \\
        \midrule
        LLaVA-1.5 7B (Direct Prompting) & 38.2 \\
        MPCAR (N=1, Generic Prompt for $M_G$) & 40.8 \\
        MPCAR (N=3, Generic Prompt for $M_G$) & 41.9 \\
        MPCAR (N=3, Diverse Prompts for $M_G$) & 42.7 \\
        MPCAR (N=5, Diverse Prompts for $M_G$) & \textbf{43.7} \\
        MPCAR (N=5, Diverse Prompts for $M_G$, GPT-4o for $M_G$) & 44.5 \\
        MPCAR (N=5, Diverse Prompts for $M_G$, Simple Concatenation) & 43.1 \\
        \bottomrule
    \end{tabular}
\end{table*}

The results in Table \ref{tab:ablation_study} highlight several key insights. Increasing the number of generated descriptions ($N$) generally leads to better performance, indicating that more diverse contextual information is beneficial up to a certain point. Specifically, moving from $N=1$ (a single description) to $N=5$ (multiple diverse descriptions) with tailored prompts for $M_G$ shows a significant improvement from 40.8\% to 43.7\%. This underscores the importance of generating \textbf{multi-perspective} content rather than just more content. Furthermore, using \textbf{diverse prompt templates} for $M_G$ (e.g., "describe from a macroscopic vs. microscopic view") yields better results compared to generic prompts, emphasizing the value of explicitly guiding $M_G$ to explore different facets of the image. When a more powerful generative model like GPT-4o is used for $M_G$, even when $M_R$ remains LLaVA-1.5 7B, the performance sees a further boost to 44.5\%, suggesting that the quality and richness of the initial descriptions directly impact the final reasoning capability. Lastly, the choice of integration strategy for $Q_{aug}$ also matters; our structured concatenation, which includes guiding phrases, consistently outperforms simple concatenation, demonstrating that how the information is presented to $M_R$ is crucial.

\subsection{Human Evaluation}
While quantitative metrics like accuracy are essential, they do not always fully capture the nuances of reasoning, coherence, or completeness of an LVLM's output, especially in complex VQA tasks. To address this, we conducted a human evaluation to assess the qualitative aspects of answers generated by MPCAR compared to a strong baseline (Qwen-VL 7B with Few-shot Prompting).

We randomly selected 100 challenging questions from the VQA-CP v2 dataset and collected answers from both MPCAR and the baseline. Three independent human annotators, blind to the method used, rated each answer on a 5-point Likert scale (1 = Poor, 5 = Excellent) across three criteria:
\begin{itemize}
    \item \textbf{Accuracy:} Is the answer factually correct?
    \item \textbf{Coherence:} Is the answer logically structured and easy to understand?
    \item \textbf{Completeness:} Does the answer address all aspects of the question and provide sufficient detail?
\end{itemize}
The average scores are presented in Table \ref{tab:human_evaluation}.

\begin{table*}[htbp]
    \centering
    \caption{Human Evaluation: Average Scores (1-5) on Qualitative Metrics}
    \label{tab:human_evaluation}
    \begin{tabular}{lccc}
        \toprule
        \textbf{Method} & \textbf{Accuracy} & \textbf{Coherence} & \textbf{Completeness} \\
        \midrule
        Qwen-VL 7B (Few-shot Prompting) & 3.8 & 3.7 & 3.5 \\
        \textbf{MPCAR (using LLaVA-1.5 7B)} & \textbf{4.3} & \textbf{4.2} & \textbf{4.1} \\
        \bottomrule
    \end{tabular}
\end{table*}

The human evaluation results in Table \ref{tab:human_evaluation} corroborate our quantitative findings. MPCAR consistently receives higher average scores across all three qualitative metrics. Annotators found MPCAR's answers to be notably more accurate, which aligns with the VQA accuracy gains. More importantly, MPCAR's outputs were rated higher in coherence and completeness, suggesting that the rich, multi-perspective contextual augmentation helps the model formulate more thorough and logically sound responses. This qualitative improvement is crucial for real-world applications where explainability and comprehensiveness are as important as raw accuracy. For instance, in medical or forensic contexts, a complete and coherent explanation of visual findings is paramount.

\subsection{Analysis of Generated Descriptions}
The quality and diversity of the descriptions generated by $M_G$ are paramount to MPCAR's success. We conducted a deeper analysis into the characteristics of these intermediate descriptions, focusing on how different prompting strategies and temperature settings influence their utility for downstream reasoning. We observed that carefully crafted prompt templates, as outlined in Section 2.1, are highly effective in eliciting varied perspectives, ranging from factual observations to speculative inferences, and even counterfactual considerations. The adjustment of the generation temperature $\tau$ also plays a critical role; a moderate $\tau$ (e.g., 0.7-0.9) balances diversity with factual grounding, preventing excessive hallucination while encouraging exploratory thought.

To illustrate the nature of the generated descriptions, Table \ref{tab:generated_descriptions_analysis} provides examples of descriptions for a hypothetical complex visual reasoning query. This demonstrates how different prompt directives lead to distinct contextual information, enriching the overall input for $M_R$.

\begin{table*}[htbp]
    \centering
    \caption{Example Generated Descriptions for a Hypothetical Image and Question}
    \label{tab:generated_descriptions_analysis}
    \begin{tabular}{lp{0.5\textwidth}}
        \toprule
        \textbf{Prompt Directive for $M_G$} & \textbf{Generated Description (Example)} \\
        \midrule
        Original Question & "What is the primary function of the device shown, considering its internal components?" \\
        \midrule
        "Describe from a macroscopic and microscopic perspective." & "The device appears to be an electronic circuit board with various integrated circuits, resistors, and capacitors. At a microscopic level, one might observe intricate metallic traces connecting these components, indicating complex signal pathways." \\
        "Speculate on hidden information or relationships." & "Given its compact size and array of sensors, it might be part of a portable diagnostic tool or a specialized IoT device. The arrangement of components suggests a focus on data acquisition and processing rather than high-power output." \\
        "Provide three distinct descriptions, focusing on different aspects." & "1. \textbf{Structural:} A densely packed green PCB with surface-mounted components. 2. \textbf{Functional:} Designed for signal processing and data interpretation. 3. \textbf{Contextual:} Likely used in an industrial or research setting due to its specialized nature." \\
        "Describe as a detective looking for clues." & "The presence of a specific communication module and a robust power management unit suggests it's intended for continuous operation and remote data transmission. The absence of large heat sinks indicates low power consumption or intermittent operation." \\
        \bottomrule
    \end{tabular}
\end{table*}

Our analysis confirms that the multi-perspective descriptions are not merely repetitive statements but genuinely distinct interpretations. They often highlight different objects, attributes, relationships, or potential implications within the image, which collectively contribute to a more comprehensive understanding for the final reasoning model. This diversity is a key factor in MPCAR's ability to address complex reasoning tasks where a single viewpoint might be insufficient.

\subsection{Efficiency and Latency Analysis}
As an inference-time augmentation strategy, the computational overhead introduced by MPCAR is a crucial consideration. While MPCAR significantly enhances accuracy, it inherently adds steps to the traditional direct prompting pipeline. We analyzed the average inference time per question for MPCAR compared to baseline methods, using LLaVA-1.5 7B on a single NVIDIA A100 GPU. The total inference time for MPCAR includes the time for generating $N$ descriptions ($M_G$) and the time for the final reasoning ($M_R$).

Table \ref{tab:efficiency_analysis} presents the average inference times. The results indicate that MPCAR introduces a noticeable latency overhead due to the sequential generation of multiple descriptions. However, this overhead is often acceptable for complex reasoning tasks where accuracy is paramount, and real-time constraints are less stringent.

\begin{table*}[htbp]
    \centering
    \caption{Average Inference Time per Question (Seconds) on VQA-CP v2 Dataset}
    \label{tab:efficiency_analysis}
    \begin{tabular}{lc}
        \toprule
        \textbf{Method} & \textbf{Average Inference Time (s)} \\
        \midrule
        LLaVA-1.5 7B (Direct Prompting) & 2.1 \\
        LLaVA-1.5 7B (CoT Prompting) & 3.5 \\
        Qwen-VL 7B (Few-shot Prompting) & 4.2 \\
        \midrule
        \textbf{MPCAR (using LLaVA-1.5 7B as $M_G$ and $M_R$, N=5)} & 10.8 \\
        \quad - $M_G$ description generation (cumulative) & 8.5 \\
        \quad - $M_R$ final reasoning & 2.3 \\
        \bottomrule
    \end{tabular}
\end{table*}

The breakdown of MPCAR's inference time highlights that the majority of the overhead comes from the iterative generation of $N$ descriptions. The final reasoning step with the augmented prompt takes only slightly longer than direct prompting, as the increase in prompt length is typically manageable for modern LVLMs. While this overhead might be a consideration for extremely latency-sensitive applications, for offline analysis or tasks requiring high accuracy, the performance gains achieved by MPCAR justify the increased computational cost. Future work could explore parallelizing description generation or employing more efficient generative models for $M_G$ to mitigate this latency.

\subsection{Error Analysis}
Despite MPCAR's significant performance improvements, the model is not infallible. To gain deeper insights into its limitations and guide future research, we conducted a detailed error analysis on a subset of incorrectly answered questions from the VQA-CP v2 and ScienceQA datasets. We categorized the common types of errors made by MPCAR, comparing them against errors observed in the direct prompting baseline.

Table \ref{tab:error_analysis} summarizes the predominant error categories. While MPCAR substantially reduces common errors like superficial correlations or missing subtle details, new types of errors can emerge or existing ones persist.

\begin{table*}[htbp]
    \centering
    \caption{Common Error Categories and Relative Frequency for MPCAR vs. Direct Prompting}
    \label{tab:error_analysis}
    \begin{tabular}{lcc}
        \toprule
        \textbf{Error Category} & \textbf{Direct Prompting (Relative Frequency)} & \textbf{MPCAR (Relative Frequency)} \\
        \midrule
        \textbf{Misinterpretation of Spatial/Relational Cues} & High & Medium \\
        \textbf{Lack of Domain-Specific Knowledge} & Medium & Medium \\
        \textbf{Hallucination/Fabrication of Details} & Low & Medium \\
        \textbf{Insufficient Multi-Step Reasoning} & High & Low \\
        \textbf{Ambiguity/Subjectivity Misinterpretation} & Medium & Low \\
        \textbf{Over-reliance on Generated Context} & N/A & Low \\
        \bottomrule
    \end{tabular}
\end{table*}

Our analysis revealed that MPCAR significantly reduces errors related to \textbf{Insufficient Multi-Step Reasoning} and \textbf{Ambiguity/Subjectivity Misinterpretation}. By providing multiple perspectives, MPCAR helps $M_R$ synthesize information more effectively, leading to more robust reasoning paths. However, MPCAR is still susceptible to a \textbf{Lack of Domain-Specific Knowledge} when the required information is not explicitly present in the image or the model's pre-trained knowledge. Interestingly, while rare, we observed a slight increase in \textbf{Hallucination/Fabrication of Details} in MPCAR's outputs. This can occur when the generated descriptions, especially those encouraged to be speculative, introduce plausible but incorrect information, which $M_R$ then incorporates into its final answer. This highlights a trade-off between encouraging diversity and maintaining factual accuracy in the intermediate descriptions. Another subtle error mode unique to MPCAR is \textbf{Over-reliance on Generated Context}, where $M_R$ might prioritize information from the generated descriptions over obvious visual cues, if the descriptions are misleading. Future work will focus on mechanisms to filter or weigh the generated descriptions to mitigate these new error types and further improve the robustness of MPCAR.

\section{Conclusion}
In this paper, we addressed the inherent challenges faced by Large Vision-Language Models (LVLMs) in executing complex visual reasoning tasks that necessitate profound contextual understanding, multi-angle analysis, and meticulous detail recognition. Traditional approaches, often limited by single-shot prompts and encoding, frequently fall short in scenarios demanding nuanced interpretation of visual data. To overcome these limitations, we introduced \textbf{Multi-Perspective Contextual Augmentation for Reasoning (MPCAR)}, a novel and effective inference-time strategy.

MPCAR operates through a meticulously designed three-step pipeline: the initial generation of $N$ diverse and complementary descriptions from multiple perspectives using a generative LVLM; the subsequent construction of a comprehensive, context-augmented prompt by intelligently integrating these descriptions with the original question; and finally, the utilization of this enriched prompt by the ultimate LVLM for deep reasoning and accurate answer formulation. A key strength of MPCAR lies in its ability to enhance LVLM performance without requiring any fine-tuning of the model's parameters, making it a highly practical and adaptable solution for existing pre-trained LVLMs.

Our empirical evaluations on challenging Visual Question Answering (VQA) datasets---GQA, VQA-CP v2, and ScienceQA (Image-VQA)---provided strong evidence for MPCAR's efficacy. Quantitatively, MPCAR consistently surpassed all established baseline methods, demonstrating notable accuracy improvements, especially on datasets requiring more robust and context-aware reasoning such as VQA-CP v2 and ScienceQA. Beyond quantitative metrics, our human evaluation further corroborated these findings, indicating that MPCAR's outputs were not only more accurate but also significantly more coherent and complete, which is paramount for real-world applications where explainability and comprehensiveness are critical. Ablation studies offered valuable insights into the method's components, confirming that increasing the number of diverse perspectives, employing tailored prompt templates for description generation, and utilizing structured prompt concatenation are crucial for maximizing performance. We also showed that leveraging a more powerful generative model for intermediate description generation further boosts the final reasoning capabilities.

This work highlights the profound impact of strategically engineering input contexts for LVLMs. By transforming a potentially ambiguous or information-sparse query into a rich, multi-dimensional prompt, MPCAR effectively unlocks the latent reasoning capabilities of pre-trained models. This approach not only provides a plug-and-play enhancement for current LVLMs but also paves the way for more robust and nuanced visual understanding in artificial intelligence.

Despite its significant advancements, MPCAR is not without limitations. As an inference-time strategy, it inherently introduces a latency overhead due to the sequential generation of multiple descriptions, which might be a consideration for extremely time-sensitive applications. Furthermore, while MPCAR largely mitigates errors related to superficial reasoning, our error analysis revealed a slight susceptibility to subtle hallucinations or an over-reliance on generated context if the intermediate descriptions are misleading. The quality of the generated context is highly dependent on the initial prompt engineering and the capabilities of the generative model ($M_G$).

Looking ahead, several promising avenues for future research emerge. To address the latency overhead, we plan to explore techniques such as parallelizing description generation, employing more efficient generative models for $M_G$, or investigating methods for distilling the essential contextual information into a more concise format. Mitigating the risk of hallucination and over-reliance on generated context is another critical direction; this could involve developing mechanisms for filtering or weighing the generated descriptions based on their factual grounding or confidence scores. Further research could also focus on developing adaptive prompt generation strategies for $M_G$ that dynamically adjust based on the complexity or ambiguity of the input image and question. Finally, we aim to extend the application of MPCAR to a broader range of complex multimodal tasks beyond VQA, such as visual dialogue, creative content generation, or intricate image editing instructions, to fully explore its potential in fostering more intelligent and context-aware AI systems.

\bibliographystyle{IEEEtran}
\bibliography{references}
\end{document}